\pdfoutput=1
\documentclass[11pt]{article}

\usepackage{acl}

\usepackage{times}
\usepackage{latexsym}
\usepackage{amsfonts}

\usepackage[T1]{fontenc}
\usepackage[utf8]{inputenc}

\usepackage{graphicx}
\usepackage{subcaption}

\usepackage{microtype}
\usepackage{cinzel} %inserted by steven to add MERLOT font

\usepackage{multirow}
\usepackage{tabularx}

\usepackage{comment} %note: inserted by Steven for block commenting
\usepackage{wrapfig} %note: inserted by Steven for wrapping text around figures
\usepackage{calligra}

\usepackage{aurical}

\newcommand\blfootnote[1]{%
  \begingroup
  \renewcommand\thefootnote{}\footnote{#1}%
  \addtocounter{footnote}{-1}%
  \endgroup
}

\makeatletter
\def\thickhline{%
  \noalign{\ifnum0=`}\fi\hrule \@height \thickarrayrulewidth \futurelet
   \reserved@a\@xthickhline}
\def\@xthickhline{\ifx\reserved@a\thickhline
               \vskip\doublerulesep
               \vskip-\thickarrayrulewidth
             \fi
      \ifnum0=`{\fi}}
\makeatother

\makeatletter
\newcommand\footnoteref[1]{\protected@xdef\@thefnmark{\ref{#1}}\@footnotemark}
\makeatother

\newlength{\thickarrayrulewidth}
\newcommand*{\affmark}[1][*]{\textsuperscript{#1}}
\newcommand{\DatasetName}{\textcinzel{CHARDat}}
\newcommand{\TaskName}{\textbf{\textcinzel{CHARD}}}
\newcommand{\dimension}{\textcinzel{dim}}
\newcommand{\rf}{\textcinzel{rf}}
\newcommand{\prev}{\textcinzel{prev}}
\newcommand{\treat}{\textcinzel{treat}}
\newcommand{\retr}{\textsc{Retr}}

\usepackage{amssymb}% http://ctan.org/pkg/amssymb
\usepackage{pifont}% http://ctan.org/pkg/pifont

\usepackage{tipa}
\usepackage{makecell}
\usepackage{enumitem,kantlipsum}

\newcommand{\titlecolor}{violet}
\title{\textcinzelblack{CHARD}: {\textcolor{\titlecolor}C}linical {\textcolor{\titlecolor}H}ealth-{\textcolor{\titlecolor}A}ware {\textcolor{\titlecolor}R}easoning Across {\textcolor{\titlecolor}D}imensions for Text Generation Models} %via Template Infilling}

\author{Steven Y. Feng\affmark[1]\thanks{\quad Work done while at CMU.}, Vivek Khetan\affmark[2], Bogdan Sacaleanu\affmark[2], Anatole Gershman\affmark[3], Eduard Hovy\affmark[3] \\ \affmark[1]Stanford University, \affmark[2]Accenture Labs, SF, \affmark[3]Carnegie Mellon University \\ \texttt{syfeng@stanford.edu} \\ \texttt{\{vivek.a.khetan,bogdan.e.sacaleanu\}@accenture.com} \\ \texttt{\{anatoleg,hovy\}@cs.cmu.edu}}

\begin{document}
\maketitle

%\vspace{-5mm}
\begin{abstract}
\vspace{-1mm}
We motivate and introduce{\TaskName}: {\textcolor{\titlecolor}C}linical {\textcolor{\titlecolor}H}ealth-{\textcolor{\titlecolor}A}ware {\textcolor{\titlecolor}R}easoning across {\textcolor{\titlecolor}D}imensions, to investigate the capability of text generation models to act as implicit clinical knowledge bases and generate free-flow textual explanations about various health-related conditions
across several dimensions. We collect and present an associated dataset,{\DatasetName}, consisting of explanations about 52 health conditions across three clinical dimensions. We conduct extensive experiments using BART and T5 along with data augmentation, and perform automatic, human, and qualitative analyses. We show that while our models can perform decently,{\TaskName} is very challenging with strong potential for further exploration.
\end{abstract}

\section{Introduction}
\label{sec:intro}
\vspace{-1mm}
%\vivek{Do we want to compare againt knowledge base approach - we will need to be more clear how those are still great. }

Pretrained language models (PLM) have seen increasing popularity for NLP tasks and applications, including text generation. %This includes seq2seq models such as BART \cite{lewis-etal-2020-bart} and T5 \cite{JMLR:v21:20-074}.
%As PLMs become more commonplace,
%Recently, 
Researchers have become interested in the extent to which PLMs can: 1) act as knowledge bases, 2) reason like humans.

%Researchers have been examining whether PLMs can act as knowledge bases, and if we can extract relevant knowledge from them, e.g. \citet{jiang-etal-2021-know}.
Rather than using external databases, exposure to large amounts of data during training combined with their large number of parameters, has given PLMs the ability to store knowledge that can be extracted through effective probing strategies such as text infilling \cite{donahue-etal-2020-enabling}, prompting \cite{prompting_survey}, and QA \cite{jiang-etal-2021-know}. PLMs %have an advantage in that they
imitate a more high-level information store, allowing for greater abstractness, flexibility, and generalizability. They are also able to better exploit contextual information than simple retrieval.%simply retrieving from an external source of data.%, which is more rigid.

%Another area of literature looks at their reasoning capabilities.
Studies have also shown that as PLMs scale up, they have have emergent abilities \cite{wei_emergent}, including reasoning. %To what extent do these emergent abilities include reasoning?
There has been increasing attention on their commonsense reasoning through works like COMET \cite{bosselut-etal-2019-comet}. However, studies show that even large PLMs struggle with commonsense tasks that humans can reason through very easily \cite{talmor2019olmpics}. There are works that investigate more complicated reasoning tasks, e.g. arithmetic and symbolic reasoning \cite{chain_of_thought}. PLMs inherently have some extent of reasoning capability, and many more complex reasoning tasks are easier to carry out over abstract PLM embedding space.

\begin{table}[t]
\centering
\small%|p{3cm}|p{3cm}|p{5cm}|
%\vspace{-4mm}
\scriptsize
\begin{tabular}{ |p{2.9cm}|p{3.9cm}| }
\hline
\textbf{Template} & \textbf{Full Text with Explanation} \\
\hline
A person with \textit{Costochondritis} has a/an \textit{exercise risk factor} because/since/as \textit{\{explanation\}} & A person with Costochondritis has an exercise risk factor because \textit{costochondritis can be aggravated by any activity that places stress on your chest area.}\\
\hline
A person with \textit{gout} has a/an \textit{lose weight prevention} because/since/as \textit{\{explanation\}} & A person with gout has a lose weight prevention because \textit{losing weight can lower uric acid levels in your body and significantly reduce the chance of gout attacks.}\\
\hline
A person with \textit{rheumatoid} has a/an \textit{therapy treatment} because/since/as \textit{\{explanation\}} & A person with rheumatoid has a therapy treatment because \textit{physiotherapy helps rheumatoid patients with pain control, reducing inflammation and joint stiffness and to return to the normal activities of daily living or sports.}\\
\hline
\end{tabular}
\vspace{-2mm}
\caption{\footnotesize Examples of {\TaskName} templates with explanations (from{\DatasetName}). The human was asked to write the entire output text (not just the explanation) by infilling the template.}
\label{tab:template_examples}
\vspace{-4mm}
\end{table}

In this paper, we are interested in the intersection of these areas. Can PLMs act as knowledge bases and also reliably reason using their own knowledge? We investigate whether PLMs can learn and reason through health-related knowledge. %\vivek{Next sentence needs more clarification - can remove my work.}
Work on generation-based reasoning for health has been limited, with most prior work exploring retrieval-based methods. % e.g. \citet{healthcare_tweets}.
Generation-based reasoning is more difficult, as such a specialized domain contains esoteric information not prevalent in the PLM's training data, and involves a higher degree of specialized reasoning to handle domain-specific problems.\blfootnote{Code: \url{https://github.com/styfeng/CHARD}}

Healthcare is an important domain that deals with human lives. It is a large application area for machine learning and NLP. The need for automation in healthcare rises, as countless studies show that healthcare workers are overworked and burned out, especially recently due to the COVID-19 pandemic \cite{healthcare_burnout,doi:10.1177/1048291120974358,10.1371/journal.pone.0257840}. Further, healthcare resources will continue to be strained as the baby boomer generation ages \cite{baby_boomer_1}.

To this end, we propose{\TaskName}: {\textcolor{\titlecolor}C}linical {\textcolor{\titlecolor}H}ealth-{\textcolor{\titlecolor}A}ware {\textcolor{\titlecolor}R}easoning across {\textcolor{\titlecolor}D}imensions (\S\ref{subsec:task}). This task is designed to explore the capability of text generation models to act as implicit clinical knowledge bases and generate textual explanations about health-related conditions across several dimensions. %similar to how a doctor may answer their patient's questions. 
The ultimate goal of{\TaskName} is to eventually have a model that %can approximate a human doctor by being 
is knowledgeable and insightful across numerous clinical dimensions and reasoning pathways. %(think of a COMET for healthcare). 
For now, we focus on three relevant clinical dimensions using a template infilling approach, and %: risk factors, treatment, and prevention, using a template infilling approach. 
collect an associated dataset,{\DatasetName}, which includes information for 52 health conditions across these dimensions (\S\ref{subsec:dataset}).

We perform extensive experiments on{\DatasetName} using two SOTA seq2seq models: BART \cite{lewis-etal-2020-bart} and T5 \cite{JMLR:v21:20-074} (\S\ref{subsec:models}), with data augmentation using backtranslation \cite{sennrich-etal-2016-improving} (\S\ref{subsec:data_aug},\ref{subsec:data_aug_experiments}). %We also explore a retrieval-based baseline that uses Google Search (\S\ref{subsec:models}).
We benchmark our models through automatic, human, and qualitative analyses (\S\ref{sec:results_and_analysis}). We show that our models %perform decently and
show strong potential, but have room to improve, %in medical accuracy and informativeness,
and that{\TaskName} is highly challenging with room for additional exploration. Lastly, we discuss several potential directions for improvement (\S\ref{sec:takeways_future_directions}).%\footnote{Code and data will be released upon publication.}

%These three dimensions are well-known facts. When we ask LM to generate that, we are using it as implicit retrieval model assuming it has seen some of this data. After that, explanation is controlled generation around a given disease and dimension. See how good they are able to generalize to other unseen diseases.
%Doctor, can you explain this disease to me? System is able to generate similar explanations for a given disease and risk factor by knowing similarities between diseases and risk factors
%Frame this as a simple explanation paper where given two concepts, it's explaining the kind of relationship

%Purpose: get multiple aspects/dimensions of diseases for reasoning, e.g. closer to COMET (single model can reason through all these dimensions)
%E.g. disease + risk factor in, home remedy out
%In the end, these three dimensions = good graph, good combination

%Justify narrower scope of templates: point is to have the LMs able to do things like diagnosing based on symptoms (reasoning) without human assistance, too much flexibility for human annotators = bullshit, templates too varied = complications/difficulties for models to learn properly

% \section{Task Formulations}
% \label{sec:task}
% \input{sections/task.tex}

\vspace{-1mm}
\section{Task and Dataset}
\label{sec:dataset}
\vspace{-1mm}
\subsection{The {\TaskName} Task}
\label{subsec:task}
\vspace{-1mm}

Our task,{\TaskName}: {\textcolor{\titlecolor}C}linical {\textcolor{\titlecolor}H}ealth-{\textcolor{\titlecolor}A}ware {\textcolor{\titlecolor}R}easoning across {\textcolor{\titlecolor}D}imensions, investigates the capability of text generation models to produce clinical explanations about various health conditions across several clinical dimensions ({\dimension}). Essentially, we assess how a PLM can be used as and reason through an implicit clinical knowledge base.% that can reason through several dimensions and generate free-flow textual explanations.

%Purpose: get multiple aspects/dimensions of diseases for reasoning, e.g. closer to COMET (single model can reason through all these dimensions)

We focus on three{\dimension}: \textbf{risk factors} (RF), \textbf{treatment} (TREAT), and \textbf{prevention} (PREV), as they are important and relevant in the context of health. A risk factor refers to something that increases the chance of developing a condition. For cancer, some examples are age, family history, and smoking. Treatment refers to something that helps treat or cure a condition. For migraines, some examples are medication, stress management, and meditation. Prevention refers to strategies to stop or lower the chance of getting a condition. For diabetes, some examples are a healthy diet and regular exercise.

As an initial approach to{\TaskName}, we use a template infilling formulation, where given an input template that lays out the structure of the desired explanation, % at a high-level,
the model's goal is to generate a complete explanation of how the particular{\dimension} attribute relates to the given condition. In particular, the templates end with an \{explanation\} span that the models fill in by explaining the appropriate relationship. Some examples are in Table \ref{tab:template_examples}.

\vspace{-1mm}
\subsection{{\DatasetName} Dataset}
\label{subsec:dataset}
\vspace{-1mm}

\paragraph{Collection Process:} We collect a dataset for our task called{\DatasetName} (where DAT is short for data). We collect data across the three{\dimension} for 52 health conditions, listed in Appendix \ref{appendix:list_of_conditions}. This is a manually curated list of health conditions which range from common conditions such as migraine and acne to rare conditions such as Lyme disease and Paget–Schroetter. The conditions were also selected by volume of online activity (e.g. number of active subreddit users), %(e.g. over 60k active visitors to the Diabetes subreddit vs. 8k for Lupus),
treatable vs. chronic conditions, %(e.g. Gout is treatable whereas Narcolepsy is chronic),
and whether a condition can be self-diagnosed or not. %(e.g. Bulimia can be self-diagnosable but Gastroparesis requires a professional).
This allows us to assess{\TaskName} across a variety of conditions. %of differing rarities and other factors. 

%\vivek{we will ned to Change the diabetes and other disease name. The screenshot are from other papaer - under review at EMNLP. That paper only uses 24 condition but we went through 50- and this green is yuck}
%\steven{i don't have time to possibly go through subreddits or search up which disease can be self-diagnosed or not, so i'm just going to remove the specific disease examples}

For each{\dimension}, we manually collect an exhaustive list of{\dimension}-related attributes (e.g. risk factors) for each condition. By \textit{attribute}, we refer to a particular example of that {\dimension} (e.g. "obesity"). This was accomplished by searching through reliable and reputable medically-reviewed sources %health resources and websites
such as MayoClinic, CDC, WebMD, and Healthline.%\footnote{\url{mayoclinic.org,cdc.gov,webmd.com,healthline.com}}.

We collect the final text (with explanations) using Amazon Mechanical Turk (AMT). We ask approved AMT workers (with strong qualifications and approval ratings on healthcare-related tasks) to write factually accurate, informative, and relatively concise passages given a particular condition and{\dimension} attribute template (per HIT), while encouraging them to consult the aforementioned health resources. Three separate annotation studies (one per{\dimension}) with strict quality control were conducted to collect an annotation per example.\footnote{Explanations for{\TaskName} are typically quite standardized, and additional annotations were repetitive. Differences are mainly in language, so we instead opt for paraphrasing data augmentation techniques such as backtranslation (\S\ref{subsec:data_aug}).} Annotations were regularly verified by authors, and a large subset of{\DatasetName} was manually examined for medical accuracy. More details are in Appendix \ref{appendix:human_dataset_details}. Some examples from{\DatasetName} are in Table \ref{tab:template_examples}. %, with more in Appendix \ref{appendix:more_dataset_examples}.

\vspace{-1mm}
\paragraph{Splits and Statistics:}%\mbox{}\\
We split{\DatasetName} by{\dimension} into train, val, and test splits of $\approx$ 70\%/15\%/15\%, and combine the individual splits per{\dimension} to form the final splits called{\DatasetName}$_{tr}$,{\DatasetName}$_{val}$, and{\DatasetName}$_{test}$, respectively. The individual{\dimension} splits are called{\dimension}$_{tr}$,{\dimension}$_{val}$, and{\dimension}$_{test}$, where{\dimension} is a short-form of the particular dimension: {\rf}, {\treat}, or {\prev}. The individual dimension subsets of {\DatasetName} are called {\DatasetName}$_{DIM}$.

For each{\dimension}'s test split, we ensure that approximately half consist of examples from conditions entirely unseen during training for that{\dimension}, called{\dimension}$_{test-unseen}$. This is to assess whether the model can generalize to unseen conditions. The other half contains examples from conditions seen during training %for that {\dimension}, 
called{\dimension}$_{test-seen}$, but the specific condition and{\dimension} attribute combination was unseen. The combined halves (across{\dimension}) are called{\DatasetName}$_{test-unseen}$ and{\DatasetName}$_{test-seen}$. We do the same for the val split to ensure consistency for model selection purposes. {\DatasetName} statistics are in Table \ref{tab:dataset_stats}. 

\begin{table}[t]
\centering
\scriptsize
\small%|p{1.8cm}|p{13.9cm}|
%\resizebox{\columnwidth}{!}{
\begin{tabular}{ |p{2.5cm}|p{0.9cm}|p{0.6cm}p{1.8cm}|}
\hline
\textbf{Dataset Stats} & Train & Val & Test (seen/unseen)\\
\hline
\# conditions = 52 & 44 & 39 & 41 (37/4)\\
  {\rf} = 52 & 44 & 26 & 26 (22/4)\\
  {\treat} = 52 & 43 & 21 & 20 (16/4)\\
  {\prev} = 44 & 35 & 11 & 21 (17/4)\\
\hline
\# sentences = 937 & 655 & 141 & 141 (70/71)\\
  {\rf} = 457 & 319 & 69 & 69 (32/37)\\
  {\treat} = 297 & 207 & 45 & 45 (20/25)\\
  {\prev} = 183 & 129 & 27 & 27 (18/9)\\
\hline
Avg length = 36.2 & 37.7 & 36.1 & 35 (35.9/34.2)\\
\hline
\end{tabular}%}
\vspace{-2mm}
\caption{\footnotesize {\DatasetName} statistics. Differing \#s by{\dimension} are because there are more risk factors for most conditions, and some do not have prevention strategies. Length is in words.% including our new splits. %Dev$_{CG}$ and test$_{CG}$ refer to our new dev and test splits, while dev$_{O}$ and test$_{O}$ refer to the original ones.
}
\label{tab:dataset_stats}
\vspace{-3mm}
\end{table}
\vspace{-1mm}

\vspace{-1mm}
\section {Methodology}
\label{sec:methodology}
\vspace{-1mm}
\subsection{Models}\label{subsec:models}
\vspace{-1mm}

\paragraph{BART and T5:} We experiment using two pretrained seq2seq models: BART and T5 (both base and large versions). These are suitable for our task formulation (template infilling). T5 \cite{JMLR:v21:20-074} has strong multitask pretraining. BART \cite{lewis-etal-2020-bart} is trained to reconstruct original text from noised text (as a denoising autoencoder). We use their HuggingFace codebases. 

\begin{figure}[t]
    \centering
    \includegraphics[width=0.45\textwidth]{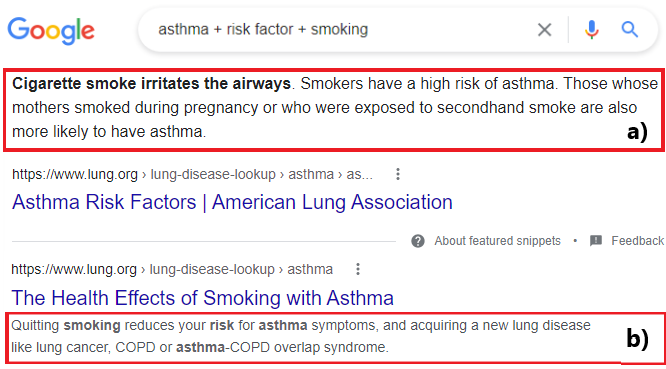}
    %\\\vspace{-0.2\abovedisplayskip}
    \vspace{-2mm}
    \caption{\footnotesize An example of the Google search results for the query \{\textit{asthma + risk factor + smoking}\} highlighting: a) the \textit{featured snippet}, b) the text below the first link.} \label{fig:google_results}
\vspace{-3mm}
\end{figure}

\vspace{-1mm}
\paragraph{Retrieval Baseline ({\retr}):} We use a retrieval-based approach as a baseline. We manually query Google using \{\textit{condition +{\dimension} +{\dimension} attribute}\}, e.g. \{\textit{asthma + risk factor + smoking}\}, and extract either the \textit{featured snippet} at the top of the results page, or the text below the first link if there is no featured snippet. If the featured snippet is a list or table, we manually concatenate the items into a single piece of text.  %with appropriate spacing and punctuation in between.
An example is in Figure \ref{fig:google_results}.

The extracted text approximates an explanation, which we then concatenate to the first part of the associated template %(while lower-casing the first letter of the retrieved explanation)
to form the final text, e.g. \textit{A person with asthma has a/an smoking risk factor because/since/as \textbf{\{retrieved explanation\}}.} {\retr} leverages Google's strong search and summarization capabilities, serving as a useful baseline. Further, Google Search is an evolving baseline that %improves over time and %further
continually challenges our{\TaskName} models.%to improve accordingly.
\footnote{We will release our current baseline data.}%, ensuring that the baseline is reproducible and consistent.}%to benchmark our trained models on{\DatasetName}.

\vspace{-1mm}
\subsection{Data Augmentation (DA)}\label{subsec:data_aug}
\vspace{-1mm}

Since {\DatasetName} is relatively small, which is mainly a function of our task and domain, i.e. there are a limited number of non-obscure medical conditions and associated{\dimension} attributes, we hypothesize that data augmentation (DA) techniques \cite{feng-etal-2021-survey,feng-etal-2020-genaug} may be useful. %for training more effective models for {\TaskName}.

As noted by \citet{feng-etal-2021-survey}, text generation and specialized domains (such as healthcare) both present several challenges for DA. In our case, many explanations contain clinical or health jargon which makes techniques that leverage lexical databases such as WordNet, e.g. synonym replacement \cite{feng-etal-2020-genaug}, challenging or impossible.

We decide to use backtranslation (BT) \cite{sennrich-etal-2016-improving} to augment examples in{\DatasetName}$_{tr}$, a popular and easy DA technique which translates a sentence into another language and back to the original language.\footnote{This is sometimes referred to as \textit{round-trip translation}.} This usually results in a slightly altered version (paraphrase) of the original text. %, as minor changes occur during the translations. 
BT is effective here as healthcare-related terms are preserved relatively well, and the resulting paraphrased explanation remains relatively intact.%is not drastically different. We wish for our explanations to remain relatively intact. %with slight vocabulary and syntactic modifications.

We use UDA \cite{xie2019unsupervised} for BT, which translates sentences from English to French, then back to English. UDA is a DA method that uses unsupervised data through consistency training on $(x,DA(x))$ pairs. An advantage of UDA's BT is that we can control for the degree of variation %compared to the original text
using a \textit{temperature (tmp)} parameter, where higher values (e.g. 0.9) result in more varied paraphrases. We only backtranslate the explanation portion of examples (concatenating them back to the preceding part) as we wish to keep the preceding part intact.

\begin{table}[t]
\centering
\small%|p{3cm}|p{3cm}|p{5cm}|
%\vspace{-4mm}
\scriptsize
\begin{tabular}{ |p{0.3cm}|p{6.8cm}| }
\hline
\textbf{Tmp} & \textbf{Text}\\
\hline
0 & A person with acne has an avoid irritants prevention because \textit{using oily or irritating personal care products clog your pores causing acne.}\\
\hline
0.5 & %A person with acne has an avoid irritants prevention because
\textit{if you use oily or irritant personal care products, you block pores and cause acne.}\\
\hline
0.7 & %A person with acne has an avoid irritants prevention because
\textit{using oily or irritating personal care products, you block %your
acne pores.}\\
\hline
0.9 & %A person with acne has an avoid irritants prevention because
\textit{use oily and irritating disinfectant products freezing your pores to cause the Acne restructurs.}\\
\thickhline
0 & A person with MultipleSclerosis has a stress management prevention because \textit{stress is more likely to exacerbate the symptoms of MS and bring about a flare or relapse.}\\
\hline
0.5 & %A person with acne has an avoid irritants prevention because
\textit{stress is more likely to exacerbate MS symptoms and lead to an outbreak or relapse}\\
\hline
0.7 & %A person with acne has an avoid irritants prevention because
\textit{stress is more likely to exacerbate symptoms of MS and trigger a flare or relapse.}\\
\hline
0.9 & %A person with acne has an avoid irritants prevention because
\textit{severe mourning problems occurred at Vancouver Hospital (Prince Edward Island), British Columbia. (...) }\\%They occurred at Children’s Hospital, Vancouver, in the province of Nunavut.}\\
\hline
\end{tabular}
\vspace{-2mm}
\caption{\footnotesize Examples of original (tmp=0) and BT text. The explanation portion (which is backtranslated) is italicized.}
\label{tab:temp_backtranslation_examples}
\vspace{-1mm}
\end{table}

\begin{figure}[t]
    \centering
    \includegraphics[width=0.45\textwidth]{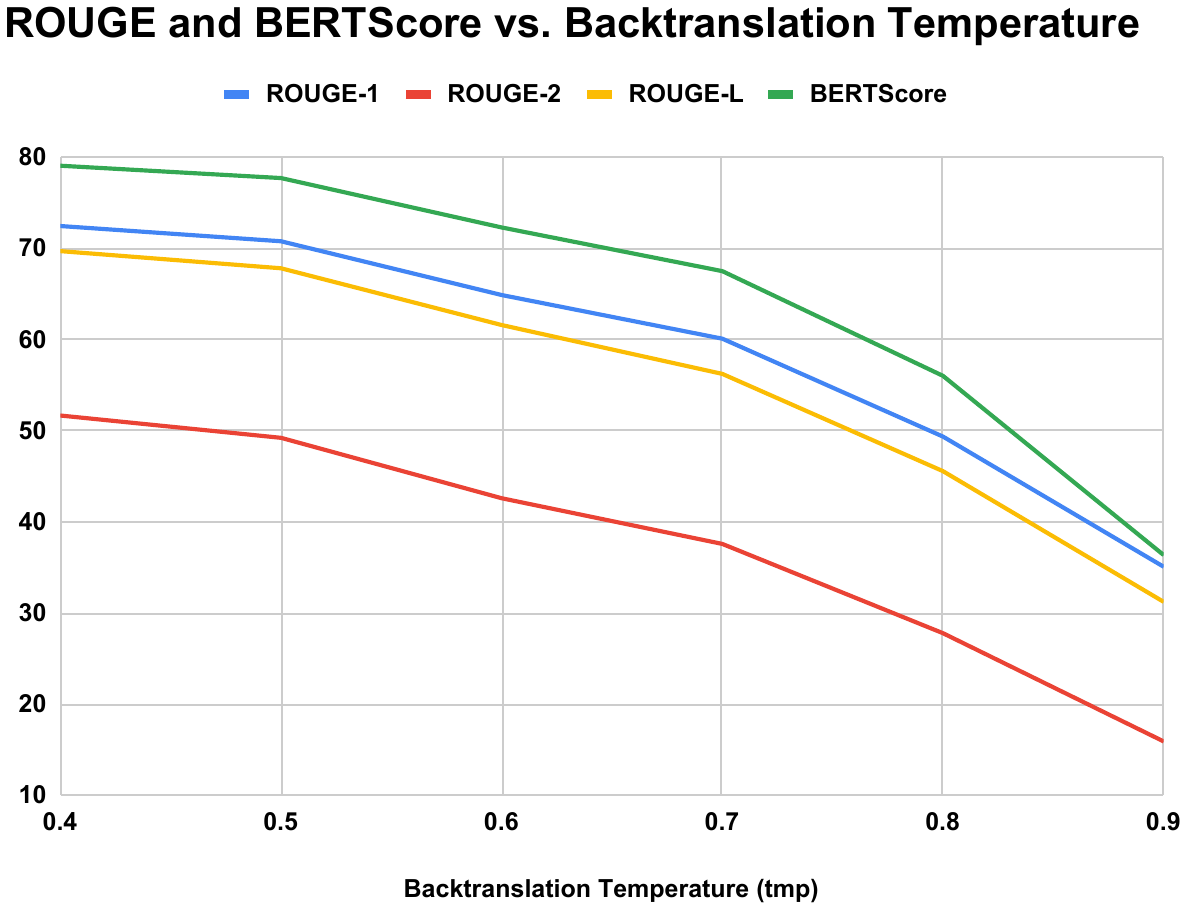}
    %\\\vspace{-0.2\abovedisplayskip}
    \vspace{-2mm}
    \caption{\footnotesize Graph showing how avg. ROUGE and BERTScore of BT vs. original text vary by BT tmp on{\DatasetName}$_{tr}$.} \label{fig:temp_backtranslation_graph}
\vspace{-3mm}
\end{figure}

From the examples in Table \ref{tab:temp_backtranslation_examples}, we can see that higher tmp typically results in more varied text, albeit with issues with content preservation and fluency. For the second example, the tmp=0.9 BT is completely unrelated to the original text. This is not entirely undesirable, as some noise may make our trained models more robust. From Figure \ref{fig:temp_backtranslation_graph}, we see that the average ROUGE and BERTScore of backtranslated{\DatasetName}$_{tr}$ text compared to the original text decrease as tmp increases, as expected.

\vspace{-1mm}
\subsection{Evaluation Metrics}\label{subsec:eval_metrics}
\vspace{-1mm}

We use several standard text generation evaluation metrics including reference-based token and semantic comparison metrics used in works like \citet{lin-etal-2020-commongen} such as ROUGE \cite{lin2003automatic},
%BLEU \cite{papineni2002bleu}, %METEOR \cite{lavie2007meteor}, C
CIDEr \cite{vedantam2015cider}, and SPICE \cite{anderson2016spice}. SPICE translates text to semantic scene graphs and calculates an F-score over graph tuples. CIDEr captures sentence similarity, grammaticality, saliency, importance, and accuracy.\footnote{Matching metrics are sufficient as{\TaskName} explanations are standardized (space for explanations is low) since our inputs present a particular condition and{\dimension} attribute combo.}

%\citet{lin-etal-2020-commongen} note that SPICE correlates highest with human judgment for CommonGen. Cov measures the average percentage of input concepts covered by the output text in any form.% \sy{NEED CITES FOR METRICS} \vg{Done all except Cov} \vg{Perhaps we should list only a couple here - anyway the reader knows which all by seeing the table.. most NLG researchers would feel drowsy seeing the same names and cites again and again lol}
 
We also use average word length (Len), BERTScore \cite{zhang2019bertscore}, and Perplexity (PPL). BERTScore serves as a more semantic similarity measure by assessing BERT \cite{devlin-etal-2019-bert} embeddings similarity between individual tokens. We multiply by 100 when reporting BERTScore. PPL approximately measures fluency, where lower values represent higher fluency. %It is defined as: $PPL(S)=exp(-\frac{1}{|S|}ln(p_M(S)))$, where $S$ is a text and $p_M(S)$ is the probability assigned to $S$ by a language model.
We use GPT-2 \cite{radford2019language} for PPL. % as it is a popular language model trained on large amounts of text data.
Higher is better for all metrics other than PPL and Len.

% \section{Categorization}
% \label{sec:categorization}
% \input{sections/categorization.tex}

% \section{Models}
% \label{sec:models}
% \input{sections/models.tex}

\vspace{-1mm}
\section{Experimental Setup}
\label{sec:experimental_setup}
\vspace{-1mm}
\subsection{Model Finetuning and Generation}
\label{subsec:model_finetuning_generation}
\vspace{-1mm}

For the standard (non-augmented){\TaskName} models, we train and evaluate four versions of each on{\DatasetName},{\DatasetName}$_{RF}$,{\DatasetName}$_{TREAT}$, and{\DatasetName}$_{PREV}$, respectively. The first of these is a combined model that learns to handle all three{\dimension} at once depending on the{\dimension} given at inference, while the latter three are models trained on each individual{\dimension}. We predict that while the latter three may perform better on their particular{\dimension}, the first model is more effective overall as it accomplishes our goal of having a single PLM that can store knowledge and reason through several{\dimension}. It is thus more adaptable and generalizable.%, and efficient for potential real-world use cases.

For training the{\TaskName} models, we keep most hyperparameters static, other than learning rate (LR) which is tuned per individual model. For each model, we select the epoch that corresponds to highest ROUGE-2 on{\DatasetName}$_{val}$, and decode using beam search. See Appendix \ref{appendix:model_finetuning_generation} for more. %details regarding hyperparameters and so forth. 

\vspace{-1mm}
\subsection{Data Augmentation Experiments}
\label{subsec:data_aug_experiments}
\vspace{-1mm}

We try several backtranslation DA experiments. 
\vspace{-1mm}

\paragraph{2x DA with Different Tmp:} Our first set of experiments involves 2x DA (backtranslating each{\DatasetName}$_{tr}$ explanation once, to 2x the original training data) using different BT tmp which we call BT-set: \{0.4, 0.5, 0.6, 0.7, 0.8, 0.9\}. We predict that the optimal tmp lies in the 0.6-0.7 range, as the text is modified to a reasonable degree.%lower tmp modify the text too little, whereas higher tmp modify too heavily.

\vspace{-1mm}
\paragraph{Different DA Amounts (2x-10x):} We also try further DA amounts: 3x, 4x, 5x, 7x, and 10x the original amount of training data. We explore whether the amount of augmentation affects performance, and hypothesize that performance will increase to a certain point and decline afterward. This is because the advantages of DA may taper off since the augmented data are variations of the original, and models may overfit past a point.

\vspace{-1mm}
\paragraph{DA Amount Strategies (best-tmp vs. diff-tmp):} We also investigate two strategies for selecting each successive iteration of augmented examples. The first is best-tmp, where all the augmented data comes from BT of the tmp that performed best for 2x DA (e.g. all from 0.7).\footnote{This is possible because UDA uses sampling, so even for the same tmp, the backtranslations differ each time.}

The second is diff-tmp, where each successive iteration is the tmp that performed next best (e.g. 2x is the best tmp, 3x is additionally the second-best tmp, etc.). For the highest DA amounts (e.g. 10x), when the six tmp values in BT-set have been exhausted, we go back to the best tmp and repeat.%(using UDA sampling) and repeat this process.

\vspace{-1mm}
\paragraph{Base vs. Large Models:} For the base models (BART-base and T5-base), we try all aforementioned tmp, DA amounts, and amount strategies. For the large models, we try the top three temperatures (for 2x DA) and amount strategy that performed best on the corresponding base model, and only 3x, 5x, 7x, and 10x DA amounts.

Note that BT tmp and DA amounts are both hyperparameters, so while we train models corresponding to different values of them, the final chosen models correspond to the ones that performed best on{\DatasetName}$_{val}$. We then use these final models to generate on{\DatasetName}$_{test}$. We report the results of the overall best models in \S\ref{sec:results_and_analysis}.

\vspace{-1mm}
\subsection{Human Evaluation}
\label{subsec:human_eval_body}
\vspace{-1mm}

We conduct human evaluation using AMT.\footnote{See Appendix \ref{appendix:human_eval_details} for further human evaluation details.} We ask two approved annotators (with strong qualifications and approval ratings on healthcare-related tasks) to each evaluate all 141{\DatasetName}$_{test}$ examples. Our evaluation uses pairwise comparison of the outputs from two methods, split into three studies per {\dimension}: {\retr} vs. best{\TaskName} model, {\retr} vs. human, and human vs. best{\TaskName} model.

We ask annotators to choose which amongst the two outputs (presented in a random order per example) has better 1) medical accuracy (MedAcc), 2) informativeness (Inform), and 3) readability (Read). Medical accuracy refers to which explanation is more clinically correct for the given{\dimension} attribute and condition. Informativeness refers to which is more complete and explains in sufficient detail (including \textit{why?}). Readability refers to which is more readable, which includes fluency (natural-sounding English) % with good grammar)
and conciseness/brevity (not overly long).% and complicated).

There are 3 choices for each evaluation aspect - O1: first is better, O2: second is better, O3: both are indistinguishable. To aggregate multiple annotations per example, we find the overall fraction of responses towards each outcome value. % as the per-example distribution. We then find the sample mean of this outcome distribution over all examples. 
%For sample mean and significance testing, we are interested in the values for O1 vs. O2.% including annotation quality control.

\vspace{-1mm}
\section{Results and Analysis}
\label{sec:results_and_analysis}
\vspace{-1mm}
\vspace{-1mm}
We report automatic results on{\DatasetName}$_{test}$ of the best models (for BART-base, BART-large, T5-base, T5-large) trained on{\DatasetName} compared to {\retr} in Table \ref{tab:auto-results-overall}. The best models are tmp=0.9 2x DA for BART (base and large), 5x DA with diff-tmp for T5-base, and tmp=0.6 2x DA for T5-large.

Our best overall{\TaskName} model is T5-large based on automatic results and qualitative analysis. We break down results of T5-large on{\DatasetName}$_{test-seen}$ and{\DatasetName}$_{test-unseen}$ in Table \ref{tab:auto-results-halves}. We show results of T5-large compared to T5-large$_{DIM}$ (models trained on the individual{\dimension}) in Table \ref{tab:auto-results-dimensions}. We conduct human evaluation with T5-large, and the results are in Table \ref{tab:human-results-overall}.% and results broken down by {\dimension} in Appendix \ref{appendix:human_eval_tables}.

\begin{figure}[t]
    \centering
    \includegraphics[width=0.45\textwidth]{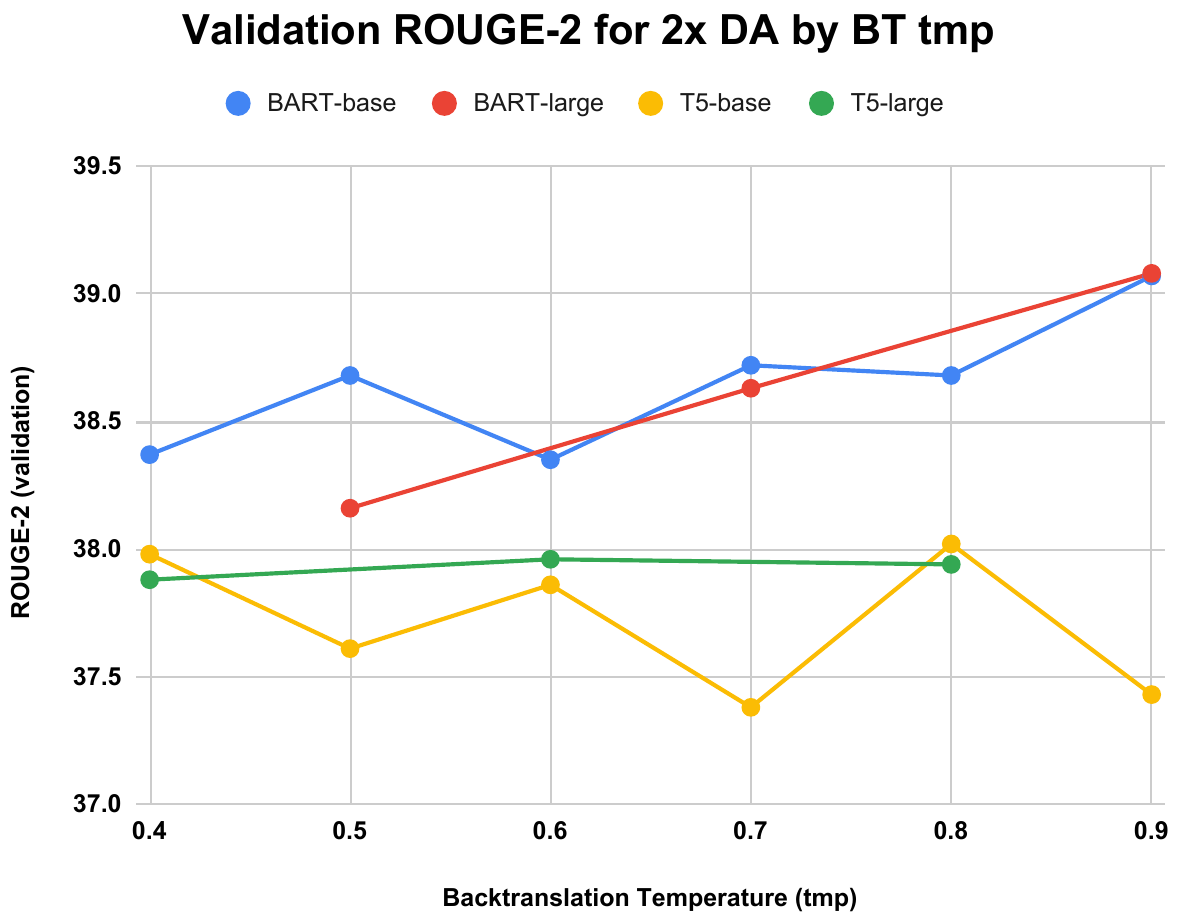}
    %\\\vspace{-0.2\abovedisplayskip}
    \vspace{-2mm}
    \caption{\footnotesize Graph showing how avg. ROUGE-2 on{\DatasetName}$_{val}$ varies by backtranslation temperature for 2x DA.} \label{fig:DA_tmp_graph}
\vspace{-1mm}
\end{figure}

\begin{figure}[t]
    \centering
    \includegraphics[width=0.45\textwidth]{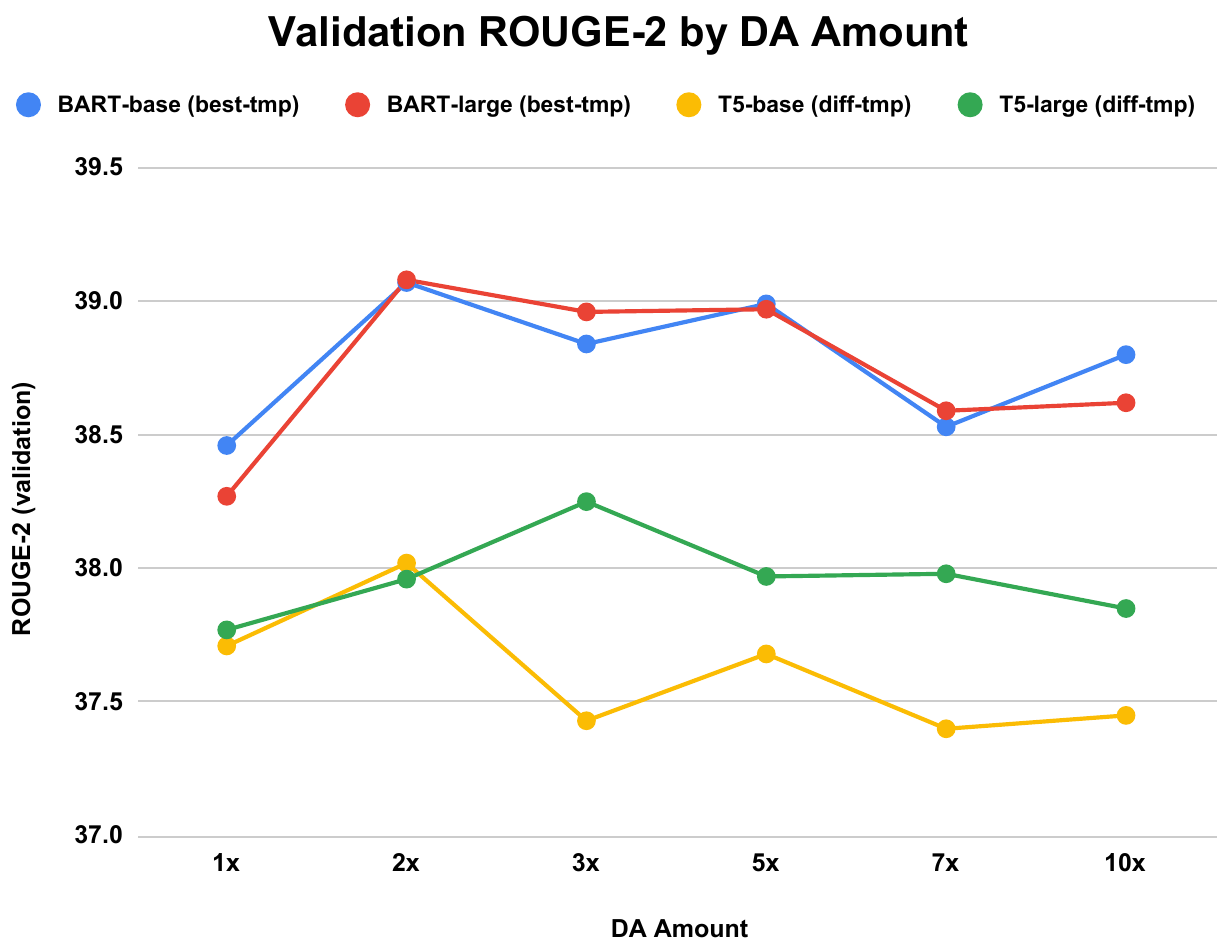}
    %\\\vspace{-0.2\abovedisplayskip}
    \vspace{-2mm}
    \caption{\footnotesize Graph showing how avg. ROUGE-2 on{\DatasetName}$_{val}$ varies by DA amount. 1x essentially refers to no DA.} \label{fig:DA_amount_graph}
\vspace{-1mm}
\end{figure}

Graphs displaying models' ROUGE-2 on{\DatasetName}$_{val}$ for 2x DA across various BT tmp and different DA amounts can be found in Figures \ref{fig:DA_tmp_graph} and \ref{fig:DA_amount_graph}, respectively. Tables \ref{tab:qualitative_body_seen} and \ref{tab:qualitative_body_unseen} contain qualitative examples, with more in Appendix \ref{appendix:qual-examples}.

\begin{table}[t]
\centering
\small
\resizebox{\columnwidth}{!}{
\begin{tabular}{ |c|c|c|c|c|c| }
\hline
 %& \multicolumn{3}{c|}{\textbf{BART-base} ($NTC=5$)} & \multicolumn{3}{c|}{\textbf{BART-large} ($NTC=2$)}\\
 %\hline
 \underline{\textbf{Metric}} & \textbf{${\retr}$} & \textbf{BART-base} & \textbf{BART-large} & \textbf{T5-base} & \textbf{T5-large}\\
 \hline
 ROUGE-1 & 43.30 & 51.37 & \textbf{51.54} & 50.00 & 50.66\\
 \hline 
 ROUGE-2 & 28.18 & 39.35 & \textbf{40.27} & 38.31 & 37.74\\
 \hline
 ROUGE-L & 39.03 & 49.55 & \textbf{49.88} & 48.07 & 48.05\\
 \hline
 BLEU-1 & 32.20 & 31.20 & 28.40 & 32.60 & \textbf{34.30}\\
 \hline
 BLEU-2 & 25.20 & 27.10 & 24.90 & 28.10 & \textbf{29.20}\\
 \hline
 BLEU-3 & 21.50 & 24.70 & 22.90 & 25.50 & \textbf{26.40}\\
 \hline
 BLEU-4 & 18.50 & 23.00 & 21.30 & 23.60 & \textbf{24.30}\\
 \hline
 METEOR & \textbf{24.40} & 22.10 & 22.10 & 21.80 & 22.10\\
 \hline
 CIDEr & 2.36 & 8.56 & 6.90 & 8.71 & \textbf{9.03}\\
 \hline
 SPICE & 35.10 & 50.50 & \textbf{50.70} & 49.10 & 49.20\\
 \hline
 BERTScore & 39.54 & 60.04 & \textbf{60.78} & 59.80 & 59.00\\
 \hline
 PPL & 65.27 & 61.00 & 87.45 & 56.78 & \textbf{52.52}\\
 \hline
 Len & 52.80 & 20.16 & 18.60 & 21.35 & 22.23\\
\hline
\end{tabular}}
\vspace{-2mm}
\caption{\footnotesize Avg. auto eval results of {\retr} and the best models (for BART and T5) on{\DatasetName}$_{test}$. Bold corresponds to best performance. For human text, PPL = 67.86, Len = 35.04.}
\vspace{-1mm}
\label{tab:auto-results-overall}
\end{table}

\begin{table}[t]
\centering
%\small
\scriptsize
%\resizebox{\columnwidth}{!}{
\begin{tabular}{ |c|c|c|c| }
\hline
 %& \multicolumn{3}{c|}{\textbf{BART-base} ($NTC=5$)} & \multicolumn{3}{c|}{\textbf{BART-large} ($NTC=2$)}\\
 %\hline
 \underline{\textbf{Metric}} & \textbf{test split (full)} & \textbf{test-seen} & \textbf{test-unseen}\\
 \hline
 ROUGE-1 & 50.66 & 49.42 & \textbf{51.93}\\
 \hline 
 ROUGE-2 & 37.74 & 37.04 & \textbf{38.35}\\
 \hline
 ROUGE-L & 48.05 & 46.98 & \textbf{49.12}\\
 \hline
 BLEU-1 & 34.30 & 33.50 & \textbf{35.20}\\
 \hline
 BLEU-2 & 29.20 & 28.60 & \textbf{29.90}\\
 \hline
 BLEU-3 & 26.40 & 25.90 & \textbf{27.00}\\
 \hline
 BLEU-4 & 24.30 & 23.80 & \textbf{24.80}\\
 \hline
 METEOR & 22.10 & 21.60 & \textbf{22.60}\\
 \hline
 CIDEr & 9.03 & \textbf{10.31} & 7.59\\
 \hline
 SPICE & 49.20 & 48.60 & \textbf{49.80}\\
 \hline
 BERTScore & 59.00 & 57.79 & \textbf{60.18}\\
 \hline
 PPL & 52.52 & \textbf{51.06} & 53.96\\
 \hline
 Len & 22.23 & 22.73 & 21.73\\
\hline
\end{tabular}%}
\vspace{-2mm}
\caption{\footnotesize Avg. auto eval results of T5-large on{\DatasetName}$_{test}$ and the test-seen and test-unseen halves.} %Bold corresponds to best performance.}
\vspace{-2mm}
\label{tab:auto-results-halves}
\end{table}

\begin{table*}[t]
\centering
%\small
\scriptsize
%\resizebox{\columnwidth}{!}{
\begin{tabular}{ |c|c|c|c|c|c|c| }
\hline %${x_n}_i$
 & \multicolumn{2}{c|}{\textbf{Risk Factors} (RF$_{test}$)} & \multicolumn{2}{c|}{\textbf{Treatment} (TREAT$_{test}$)} & \multicolumn{2}{c|}{\textbf{Prevention} (PREV$_{test}$)}\\
 \hline
 \underline{\textbf{Metric}} & \textbf{T5-large} & \textbf{T5-large$_{RF}$} & \textbf{T5-large} & \textbf{T5-large$_{TREAT}$} & \textbf{T5-large} & \textbf{T5-large$_{PREV}$}\\
 \hline
 ROUGE-1 & 52.74 & \textbf{53.17} & \textbf{49.42} & 47.38 & 47.73 & \textbf{50.00}\\
 \hline 
 ROUGE-2 & 40.52 & \textbf{41.88} & 36.12 & \textbf{36.69} & 33.00 & \textbf{36.10}\\
 \hline
 ROUGE-L & 50.40 & \textbf{51.03} & \textbf{46.60} & 45.54 & 44.43 & \textbf{48.19}\\
 \hline
 BLEU-1 & \textbf{34.80} & 34.70 & \textbf{31.10} & 25.70 & \textbf{30.80} & 29.80\\
 \hline
 BLEU-2 & 30.40 & \textbf{30.90} & \textbf{26.30} & 22.40 & \textbf{25.10} & 25.00\\
 \hline
 BLEU-3 & 27.90 & \textbf{28.60} & \textbf{23.90} & 20.70 & 21.90 & \textbf{22.00}\\
 \hline
 BLEU-4 & 26.10 & \textbf{26.90} & \textbf{22.10} & 19.30 & \textbf{19.30} & \textbf{19.30}\\
 \hline
 METEOR & 23.00 & \textbf{24.20} & \textbf{20.70} & 19.20 & 20.50 & \textbf{20.80}\\
 \hline
 CIDEr & \textbf{13.50} & 10.57 & 5.06 & \textbf{5.98} & \textbf{5.88} & 5.83\\
 \hline
 SPICE & 49.90 & \textbf{51.50} & \textbf{46.60} & 45.30 & 46.50 & \textbf{46.60}\\
 \hline
 BERTScore & 60.40 & \textbf{61.03} & \textbf{58.07} & 56.60 & 56.90 & \textbf{59.09}\\
 \hline
 PPL & \textbf{40.90} & 58.92 & \textbf{52.13} & 86.15 & \textbf{84.06} & 110.66\\
 \hline
 Len & 22.30 & 20.28 & 22.09 & 19.82 & 22.27 & 20.52\\
\hline
\end{tabular}%}
\vspace{-2mm}
\caption{\footnotesize Breakdown of the avg. auto eval results of T5-large compared to T5-large$_{DIM}$ models (trained on the three individual{\dimension}) on the respective{\dimension} subsets of{\DatasetName}$_{test}$. Bold corresponds to best performance per{\dimension}.}
\vspace{-2mm}
\label{tab:auto-results-dimensions}
\end{table*}

%\begin{wrapfigure}{r}{0.55\textwidth}
\begin{table}[!ht]
\centering
\scriptsize
%\resizebox{\columnwidth}{!}{
%\vspace{-4mm}
\begin{tabular}{ |c|c|c|c|c| }
\hline
\underline{Methods} & \underline{Aspect} & \textbf{O1} & \textbf{O2} & \textbf{O3}\\
 \hline
 \multirow{3}{*}{\textbf{{\retr} vs. Human}} & MedAcc & 0.45 & \textbf{0.53} & 0.02\\
 \cline{2-5}
 & Inform & 0.45 & \textbf{0.53} & 0.02\\
 \cline{2-5}
 & Read & 0.22 & \textbf{0.69} & 0.09\\
 \hline
 \multirow{3}{*}{\textbf{Human vs. T5}} & MedAcc & \textbf{0.72} & 0.24 & 0.04\\
 \cline{2-5}
 & Inform & \textbf{0.72} & 0.25 & 0.03\\
 \cline{2-5}
 & Read & 0.41 & \textbf{0.49} & 0.10\\
 \hline
 \multirow{3}{*}{\textbf{{\retr} vs. T5}} & MedAcc & \textbf{0.73} & 0.25 & 0.02\\
 \cline{2-5}
 & Inform & \textbf{0.73} & 0.26 & 0.01\\
 \cline{2-5}
 & Read & 0.35 & \textbf{0.62} & 0.03\\
 \hline
\end{tabular}
\vspace{-2mm}
\captionof{table}{\footnotesize Avg. human eval results on{\DatasetName}$_{test}$. O1: first method wins, O2: second method wins, O3: indistinguishable. Bold corresponds to higher fractional outcome. T5 refers to T5-large. % between O1 and O2 per aspect. 
See \S\ref{subsec:human_eval_body} and Appendix \ref{appendix:human_eval_details} for further details.}
\label{tab:human-results-overall}
\vspace{-3mm}
\end{table}
%\end{wrapfigure}

\begin{table}[h!]
\centering
\small
%\scriptsize
%\begin{center}
\scriptsize
\addtolength{\tabcolsep}{-4pt}
\resizebox{\columnwidth}{!}{
\begin{tabular}{|p{0.9cm}|p{8.2cm}|}
\hline 
\textbf{Method} & \textbf{Text} \\ \hline
Input (1) & A person with Hyperhidrosis has a medication treatment because \{explanation\}\\\hline
Human & {\textcolor{teal}{A person with Hyperhidrosis has a medication treatment because anticholinergics medicines such as glycopyrrolate, oxybutynin, benztropine, propantheline etc are helping to prevent the stimulation of sweat glands and, thus, decreasing sweating.}}\\\hline
{\retr} & \textcolor{red}{A person with Hyperhidrosis has a medication treatment because the most common medications used to manage excessive sweating are anticholinergics, including but not limited to: Glycopyrrolate (Robinul). Oxybutynin (Ditropan). Benztropine. Propantheline bromide.} \\ \hline
T5-L & {\textcolor{blue}{A person with Hyperhidrosis has a medication treatment because medications such as ibuprofen, phenytoin, and phenobarbital can help reduce the amount of sweating in the body.}} \\ \hline 
T5-B & {\textcolor{brown}{A person with Hyperhidrosis has a medication treatment because it helps to reduce the amount of blood in the body and reduces the risk of a blood clot.}} \\
\Xhline{3\arrayrulewidth}
Input (2) & A person with Alzheimers has an engage in activity prevention because \{explanation\}\\\hline
Human & {\textcolor{teal}{A person with Alzheimers has an engage in activity prevention because engaging in activity will increase nerve connections in the brain or release hormones that increase nerve cell creation and this helps preserve nerve cells in the brain and mental function}}\\\hline
%{\retr} & \textcolor{red}{A person with Alzheimers has a/an engage in activity prevention because/since/as physical activity has many health benefits, such as reducing falls, maintaining mobility and independence, and reducing the risk of chronic ...} \\ \hline
T5-L & {\textcolor{blue}{A person with Alzheimers has an engage in activity prevention because activities such as walking, jogging, swimming, and playing games can help prevent Alzheimer's}} \\ \hline 
BART-B & {\textcolor{orange}{A person with Alzheimers has an engage in activity prevention because a person with Alzheimer's disease is more likely to develop dementia}} \\
\Xhline{3\arrayrulewidth}
Input (3) & A person with Prostatitis has an infection risk factor because \{explanation\}\\\hline
Human & {\textcolor{teal}{A person with Prostatitis has an infection risk factor because the condition is caused by common strains of bacteria which spread from other parts of the urinary or reproductive systems to infect and inflame the prostate resulting in prostatitis.}}\\\hline
%{\retr} & \textcolor{red}{A person with Prostatitis has a/an infection risk factor because/since/as risk factors. Infection of the urinary or reproductive system. HIV infection or AIDS. Use of a tube inserted into the urethra to drain the bladder (urinary catheter) Diagnostic sampling of prostate tissue (biopsy)} \\ \hline
T5-L & {\textcolor{blue}{A person with Prostatitis has an infection risk factor because infections of the prostate, urethra, and genital tract can lead to prostatitis.}} \\ \hline 
%BART-large & {\textcolor{violet}{A person with Prostatitis has an infection risk factor because infection can cause inflammation of the prostate.}} \\
\end{tabular}}
\vspace{-2mm}
\caption{\footnotesize Qual examples for{\DatasetName}$_{test-seen}$: \textcolor{teal}{human}, \textcolor{red}{{\retr}}, \textcolor{blue}{T5-large (L)}, \textcolor{brown}{T5-base (B)}, \textcolor{orange}{BART-base (B)}.}
\label{tab:qualitative_body_seen}
\vspace{-3mm}
\end{table}
%\vspace*{-2mm}

\begin{table}[h!]
\centering
\small
%\scriptsize
%\begin{center}
\scriptsize
\addtolength{\tabcolsep}{-4pt}
\resizebox{\columnwidth}{!}{
\begin{tabular}{|p{0.9cm}|p{8.2cm}|}
\hline
\textbf{Method} & \textbf{Text} \\
\hline 
Input (4) & A person with Costochondritis has a fix improper posture prevention because \{explanation\}\\\hline
Human & {\textcolor{teal}{A person with Costochondritis has an fix improper posture prevention because bad posture puts stress on joints and tissues, causing joint stiffness, weakened muscles, and inflammation that is so common in Costochondritis.}}\\\hline
{\retr} & \textcolor{red}{A person with Costochondritis has a fix improper posture prevention because what can I do to relieve costochondritis? Lifestyle and home remedies. Nonprescription pain relievers. Acetaminophen (Tylenol, others) ibuprofen (Advil, Motrin IB, others) or naproxen sodium (Aleve, others) (...)} %may be helpful. Topical pain relievers. These include creams, gels, patches and sprays. ... Heat or ice. ... Rest.}
\\ \hline
T5-L & {\textcolor{blue}{A person with Costochondritis has a fix improper posture prevention because poor posture can lead to pain and stiffness.}} \\ %\hline 
%BART-base & {\textcolor{blue}{A person with Costochondritis has a fix improper posture prevention because a person who has a chronic pain disorder is more likely to suffer from chronic pain}} \\
\Xhline{3\arrayrulewidth}
Input (5) & A person with rheumatoid has a diet risk factor because \{explanation\}\\\hline
Human & {\textcolor{teal}{A person with rheumatoid has a diet risk factor because unhealthy food choices can increase inflammation from rheumatoid}}\\\hline
{\retr} & \textcolor{red}{A person with rheumatoid has a diet risk factor because the results of this study revealed that high intakes of carbohydrates (e.g., potatoes) might be associated with increased RA risks.} \\ \hline
T5-L & {\textcolor{blue}{A person with rheumatoid has a diet risk factor because people who eat a high-fat diet are at a higher risk of developing the disease.}} \\ \hline 
BART-L & {\textcolor{violet}{A person with rheumatoid has a diet risk factor because a diet rich in red meat, dairy products, and dairy products can cause inflammation}} \\
\Xhline{3\arrayrulewidth}
Input (6) & A person with depression has a life-changing event risk factor because \{explanation\}\\\hline
Human & {\textcolor{teal}{A person with depression has a life-changing event risk factor because approximately 70\% of first depression episodes and 40\% of recurrent episodes of depression are preceded by a severe life event}}\\\hline
{\retr} & \textcolor{red}{A person with depression has a life-changing event risk factor because psychological and Psychosocial Risk Factors for Recurrence · by SL Burcusa · 2007 · Cited by 1085 — These factors include age at onset of the first episode, lifetime number of} \\ \hline
T5-L & {\textcolor{blue}{A person with depression has a life-changing event risk factor because a major life event, such as the death of a loved one, can increase the risk of depression.}} \\ \hline 
%T5-B & {\textcolor{violet}{A person with depression has a life-changing event risk factor because traumatic events can lead to depression.}} \\\hline
\end{tabular}}
\vspace{-2mm}
\caption{\footnotesize Qual examples for{\DatasetName}$_{test-unseen}$: \textcolor{teal}{human}, \textcolor{red}{{\retr}}, \textcolor{blue}{T5-large (L)}, \textcolor{violet}{BART-large (L)}.}
\label{tab:qualitative_body_unseen}
\vspace{-3mm}
\end{table}
%\vspace*{-2mm}

\vspace{-1mm}
\subsection{Automatic Evaluation Results}
\label{subsec:auto_eval_analysis}
\vspace{-1mm}

From Table \ref{tab:auto-results-overall}, we see that all{\TaskName} models perform better than {\retr} across most metrics. {\retr}'s average outputs are much longer than those of our models and humans. Among our models, T5-large and BART-large perform best, demonstrating that larger models are more adept. T5-large performs best overall (combined with the qual analysis in \S\ref{subsec:qualitative_analysis}), with the longest average outputs among our models. Some of our models achieve better average fluency (PPL) compared to humans, but the outputs are generally noticeably shorter.

From Table \ref{tab:auto-results-halves}, we see that T5-large surprisingly performs better on the test-unseen half. It appears that the model can generalize decently to unseen conditions when trained on{\DatasetName}. This may partially be due to similar explanations for particular{\dimension} attributes across conditions, e.g. why \textit{sleep} helps treat %X or Y 
some conditions may be similar. %Further, the test-seen half contains many more different conditions, which may be another reason for lower performance on it.

From Table \ref{tab:auto-results-dimensions}, we see that for most{\dimension} (namely RF and PREV), the model trained on that specific{\dimension} performs better on that{\dimension}. However, our general T5-large model performs better on TREAT. It may be that training on{\DatasetName} has allowed the model to learn from data of other{\dimension}, improving its overall knowledge and generation capabilities (an advantage of a single combined model).%, and in this case, has helped for TREAT.% specifically.

From Figure \ref{fig:DA_tmp_graph}, we see that the BART models generally increase in performance with higher BT tmp (upward trend), whereas T5 fluctuates. This may be due several reasons, e.g. differences in the architecture and pretraining strategies of the models, allowing BART to leverage noisy data more effectively. From Figure \ref{fig:DA_amount_graph}, we see that performance generally increases for each model up to a certain point (e.g. 2x or 3x DA), and then decreases afterward, aligning with our hypothesis from \S\ref{subsec:data_aug_experiments}.

\vspace{-1mm}
\subsection{Human Evaluation Results}
\label{subsec:human_eval_analysis}
\vspace{-1mm}

From Table \ref{tab:human-results-overall}, we see that both {\retr} and T5-large are outperformed by humans, although {\retr} is relatively close in informativeness and medical accuracy, and T5-large slightly outperforms on readability. {\retr} outperforms T5-large on medical accuracy and informativeness, which is somewhat expected as it uses Google Search. It is worse than T5-large on readability, as our models generate more fluent, concise, and readable text (see \S\ref{subsec:qualitative_analysis}). %Hence, {\retr}'s low automatic evaluation scores in Table \ref{tab:auto-results-overall} may be due to its long generations that reduce readability and token + semantic matching scores when compared to {\DatasetName$_{test}$}, although they may generally be more medically accurate and informative.

\vspace{-1mm}
\subsection{Qualitative Analysis}
\label{subsec:qualitative_analysis}
\vspace{-1mm}

We examine the qualitative examples in Tables \ref{tab:qualitative_body_seen} and \ref{tab:qualitative_body_unseen}. Firstly, we see that {\retr} is able to generally perform well by extracting relevant information (ex.1 - a list of medications for Hyperhidrosis, ex.5 - that carbohydrates increase RA risk), which is expected using Google Search. However, it sometimes extracts a lengthy amount of irrelevant information. % that is not easily readable.
For ex.4, {\retr} extracts a difficult-to-read list of different TREAT strategies, which is for the wrong{\dimension}, and does not narrow down on an explanation for the specific{\dimension} attribute in the input. %PREV strategy of \textit{"fix improper posture"}.
For ex.6, it extracts the info and beginning of a passage from a scientific article, ending abruptly and not explaining the given{\dimension} attribute. %for depression at all.

Our models, specifically T5-large, are generally able to output more concise, readable, and sometimes relevant explanations compared to {\retr}. For ex.1, T5-large outputs a list of medications, albeit not for Hyperhidrosis - showing weaknesses in medical accuracy. Other than ibuprofen, the other medications are not in{\DatasetName}$_{tr}$, showing that these were likely already known to T5-large through pretraining. For ex.2, it generates a reasonable list of activities to help prevent Alzheimer's, and for ex.3, it lists correct body parts where an infection can occur to cause Prostatitis. It can generalize well to unseen conditions, as shown through ex.4-6. It reasons that poor posture can lead to pain and stiffness, high-fat diets can increase the chance of rheumatoid, and that a major life event (\textit{"death of a loved one"}) can cause depression. These generalization capabilities are likely from a combination of pretraining and{\DatasetName}$_{tr}$.% Hence, T5-large is able to reason to a fair extent to generate logically-grounded explanations, and mistakes n medical accuracy are mainly due to stating wrong facts. Hence, its weaknesses may be more heavily due to knowledge storage compared to reasoning.%and training on {\DatasetName}$_{tr}$ specifically. 

Compared to humans, T5-large's outputs are lacking. Human explanations are typically longer and more informative, explaining the exact reason (\textit{why?}) a specific{\dimension} attribute relates to the given condition. For ex.2, it explains how activities can help \textit{"preserve nerve cells in the brain and mental function"}, whereas T5-large simply lists activities. This similarly occurs for ex.3-5. Human explanations are also typically more medically accurate, e.g. for ex.1, the listed medications are correct. However, we do see that some of T5-large's outputs (for ex.1,2,4) are more readable. Further, T5-large sometimes presents more information, e.g. an exact list of activities for ex.2, a specific type of diet (\textit{"high-fat"}) for ex.5 (human just says \textit{"unhealthy"}), and an example of a life-changing event for ex.6.

BART-large also performs decently. In ex.5, it lists several specific and correct types of foods (\textit{"red meat, dairy products"}). The base models perform much worse. For ex.1, T5-base talks about medication reducing %\textit{"blood} and
\textit{"blood clots"}, unrelated to Hyperhydrosis. For ex.2, BART-base writes an explanation completely irrelevant to the input{\dimension}. %and rather talks about how Alzheimer's can lead to dementia.

\vspace{-1mm}
\section{Directions for Improvement}
\label{sec:takeways_future_directions}
\vspace{-1mm}
%we see that our large models (T5-large in particular) greatly outperform the base models and have decent generalization. 
We see that our models are decent and generate readable text, but can improve on medical accuracy and informativeness. %(specifically, explaining \textit{why?}) compared to both humans and {\retr}.  %but this is expected as {\retr} leverages Google Search, the most effective search engine and retrieval platform. However, the models usually outperform {\retr} in regards to readability, brevity, and sometimes relevancy.
While they are not nearly ready for real-world use, they show potential.% in \S\ref{subsec:qualitative_analysis}. 

As stated, the purpose of {\TaskName} is to assess the capabilities of PLMs to act as implicit clinical knowledge bases that can reason through several dimensions. How can we improve our models, and possibly our dataset and task formulation?

\vspace{-1mm}
\paragraph{Dataset and task formulation:} We introduce {\TaskName} and initially frame the task using a template infilling approach which is more constrained. %and narrows the scope of potential generations.
More flexible formulations may better leverage the knowledge and generation capabilities of PLMs.

Our current approach involves generating explanations about a single condition and{\dimension} attribute at a time. We can possibly improve{\DatasetName} by annotating for more complicated input %templates (or more generally,
queries. %may be more effective for training and assessing models for {\TaskName}.
This is because %abstractness, conciseness, and generalizability are likely more achievable with an LM (that acts like a high-level knowledge store) which is able to better exploit contextual information than a retrieval-based mechanism which is more rigid. 
a PLM may be more effective at answering more complicated queries, e.g. comparing and contrasting conditions and{\dimension} and multi-hop reasoning. It is likely easier to make complicated inferences and connections over the abstract PLM embedding space than over retrieved text passages.

Further, we can expand{\DatasetName} to include more dimensions and topics in the health domain. % within the health domain.
These improvements may allow for the training of a single system that is able to make complicated clinical inferences %and perform commonsense inferences
across various topics and{\dimension}. %resulting in a system potentially similar to COMET \cite{bosselut-etal-2019-comet} but for the health domain. 
%This would help to make progress towards the ultimate goal of {\TaskName}, which is to eventually have a language model that is capable of approximating a human doctor.

\vspace{-1mm}
\paragraph{Model improvements:}
%To improve %the medical accuracy and informativeness of
%our models,
We can explore models such as GPT-3 \cite{gpt3} and PALM \cite{palm_google} for{\TaskName} that are larger with stronger pretraining. %(since we showed in \S\ref{sec:results_and_analysis} that large models perform better).
We can also investigate enhancing PLMs with information retrieval, e.g. using a retrieval approach to obtain relevant scientific literature as evidence, %from several sources,
combined with a text summarization system to digest the content. Our model can then conduct its clinical reasoning on this digested content. Users can potentially take advantage of such a system to automatically verify the medical accuracy of generated explanations, %(i.e. claim verification),
and then improve the generation model itself using this feedback loop (i.e. a self-improving system).

\vspace{-1mm}
\section{Related Work}
\label{sec:related_work}
\vspace{-1mm}
\paragraph{Constrained Text Generation:} There have been several works on constrained text generation. %\citet{miao2019cgmh} use Metropolis-Hastings sampling to determine Levenshtein edits per generation step.
For creative text generation, \citet{NAREOR} introduce narrative reordering (NAREOR) to edit the temporality of narratives. \citet{PINEAPPLE} and \citet{PANCETTA} explore the generation of personifications and tongue twisters, respectively. %There has also been work specifically on template infilling. 
\citet{donahue-etal-2020-enabling} introduce and investigate the task of infilling. \citet{feng-etal-2019-keep} propose Semantic Text Exchange to adjust topic-level text semantics using infilling. \citet{dheeraj_paper} investigate cross-domain reasoning using a prompt-tuning setup. Our work distinctly investigates template infilling for clinical reasoning along dimensions.

\vspace{-1mm}
\paragraph{Commonsense Reasoning for Models:} %There have been several works which investigate commonsense reasoning for NLP models.
%\citet{talmor2019olmpics} show that not all PLMs can reason through commonsense tasks.
%There are works that investigate commonsense injection into models; one popular way is through knowledge graphs (KGs).
One large commonsense KG is COMET, which trains on KG edges to learn connections between words and phrases. COSMIC \cite{ghosal2020cosmic} uses COMET to inject commonsense into models. CommonGen \cite{lin-etal-2020-commongen} assesses the commonsense reasoning of text generation models. %., through generating logical text based on input keywords representing everyday concepts.
Several works investigate CommonGen, including SAPPHIRE \cite{feng-etal-2021-sapphire} and VisCTG \cite{feng_caption}, the latter of which uses visual grounding.  %\cite{liu-etal-2022-generated} investigate prompting to generate knowledge statements from LMs, and using its own generations to improve commonsense reasoning.
Unlike these works,{\TaskName} distinctly investigates reasoning for the clinical/health domain.

\vspace{-1mm}
\paragraph{Reasoning for Clinical/Health Domain:}  Most existing work here involves retrieval or extraction. MIMICause \cite{khetan-etal-2022-mimicause} extracts causal medical information from electronic health records to help understand narratives in clinical texts.
\citet{info:doi/10.2196/37201} extract a causal graph and reason about diabetes distress for better understanding the opinions, feelings, and observations of the diabetes online community from a causality perspective. %\citet{healthcare_tweets} identify causal relations in tweets involving diabetes. %social media through information extraction, specifically by looking at diabetes-related tweets. 

For generation, \citet{health_simplification} investigate the use of LMs %and word frequencies
to simplify medical text. \citet{abaho-etal-2022-position} probe factual knowledge from LMs %using a position-attention mechanism %that involves masking single tokens
to elicit answers related to treatment outcomes.%(observations or measurements to capture and assess the effect of treatments \cite{pubmed_outcomes}).
{\TaskName} has a different goal: rather than simply probe for factual knowledge, we assess how LMs can act as and reason through an implicit knowledge base. %with information across several dimensions, and reason through this knowledge with a more flexible template infilling approach. 
\citet{meng-etal-2022-rewire} investigate probing biomedical knowledge by introducing a benchmark, MedLAMA, that focuses on 19 relations.{\TaskName} instead focuses on clinical knowledge reasoning along different dimensions.

\vspace{-1mm}
\section{Conclusion and Future Work}
\label{sec:conclusion_future_work}
\vspace{-1mm}
In conclusion, we proposed and investigated the task of{\TaskName}: {\textcolor{\titlecolor}C}linical {\textcolor{\titlecolor}H}ealth-{\textcolor{\titlecolor}A}ware {\textcolor{\titlecolor}R}easoning across {\textcolor{\titlecolor}D}imensions, to explore the capability of text generation models to act as implicit clinical knowledge bases and generate explanations %about various health-related conditions
across several health dimensions. We presented a dataset,{\DatasetName}, and conducted experiments with BART and T5. %, T5, data augmentation, and a retrieval baseline.
Extensive evaluation and qualitative analysis demonstrated that our models are decent, especially for generating concise and readable text, but can be improved on medical accuracy and informativeness, and that{\TaskName} is challenging with much potential for further exploration. We highly encourage the research community to further investigate and improve upon{\TaskName}.

Future directions are discussed in \S\ref{sec:takeways_future_directions}. Some additional ideas include trying more data augmentation strategies %such as  %synthetic noise and
%Semantic Text Exchange, %\cite{feng-etal-2019-keep,feng-etal-2020-genaug},
and %different
decoding strategies for text infilling.%the use of decoding/sampling rather than beam search for decoding to potentially improve diversity when infilling templates.

%\newpage
\section*{Limitations}

We discuss some limitations of our work and potential directions for improvement in \S\ref{sec:takeways_future_directions}. Specifically, our template-infilling approach is less flexible, and we can expand to more complicated input queries to better leverage the power of PLMs in future work. Further,{\DatasetName} focuses on three main clinical dimensions, which can be expanded upon to include more dimensions and topics in the future. Our seq2seq models are also relatively weaker compared to GPT-3, PALM, and recent larger PLMs, which may perform more effectively on{\TaskName}. We are also investigating a completely generative approach, and combining generation with retrieval in interesting ways may be more effective. Overall, our current{\TaskName} models have room to improve on medical accuracy and informativeness, and are not nearly ready for real-world use. 

However, we note again that we are the first to propose{\TaskName}, and our work is the first step towards longer-term goals regarding clinical reasoning using PLMs. We are after more of the \textit{commonsense} medical reasoning for now, rather than very deep medical knowledge. In this paper, we see how far one can get with a standard task formulation, NLP methods, seq2seq models, and AMT annotations. As they say, \textit{"walk before you run"}!
\section*{Ethics}

We collected {\DatasetName} and conducted our human evaluation studies using AMT, in a manner consistent with terms of use of any sources and intellectual property and privacy rights of AMT crowd workers.%We neither solicit, record, nor request any kind of personal information from the annotators. 

Our collected dataset, {\DatasetName}, consists of general clinical information, where explanations are impersonal.
%We encouraged the AMT workers to use reliable health resources and websites such as MayoClinic, CDC, WebMD, and Healthline, which do not contain any personal or private info. Hence, our dataset itself contains no personal or private information.
We also manually examined a large subset of the data, and ensured there were no issues with respect to privacy and other ethical concerns, e.g. offensive words, profanities, racism, gender bias, and other malicious language.

%Crowd workers were fairly compensated: \$0.2 per evaluation HIT for a roughly 1 min task. This is at least 1.5-2 times the minimum wage in the U.S.A. of \$7.25 per hour (\$0.12 per minute).

    %-
    %-pay for eval: \$0.2 per hit, overall pay was higher than \$X, higher than min wage (look at Dheeraj paper)

We acknowledge the weaknesses of {\TaskName} models and the potential risks if they are used for real-world purposes. We will never use our models or encourage their use for real-world purposes, at least in their current state, and also emphasize this in the paper. As we noted, we propose {\TaskName} and conduct our initial experiments purely for investigation purposes and to test our hypotheses. Our paper presents an important contribution to the ML, NLP, and healthcare communities, and we encourage researchers to further improve upon it.

Our task, models, dataset, and accompanying publication are intended only for research purposes and to assess the capabilities of text generators.

%Other than health-specific challenges in regards to potential ethical concerns, we do not foresee any other explicit way malicious actors could specifically misuse our dataset, or models that could be trained on it, beyond the misuse that is possible in general with any transduction task or dataset (e.g summarization). 

%Since our dataset could be used to learn a paraphrasing function, all possible misuses that are conceivable for other paraphrasing transforms (e.g. syntactic paraphrasing) are also conceivable using our dataset.

%NLG models are known to suffer from biases learnable from training or finetuning on data, such as gender bias \cite{dinan2020queens}. However, our work and contribution does not present or release any completely new model architectures, and is primarily concerned with more careful adaptation and finetuning of existing pretrained models for a particular domain of knowledge (i.e. clinical and health). The frailties, vulnerabilities, and potential dangers of these models have been well researched and documented, and a specific re-investigation would be repetitive and beyond the scope and space constraints of this paper.

%\clearpage
%\newpage
%\input{sections/limitations}

%
%\section*{Acknowledgements}

\bibliography{anthology,custom}
\bibliographystyle{acl_natbib}

\clearpage
\appendix
% \section{Appendices}
% \label{sec:appendix}
\section{Full List of Health Conditions}
\label{appendix:list_of_conditions}

See Table \ref{tab:all_health_conditions} for a list of all health conditions in {\DatasetName}.

\begin{table*}[t]
\centering
\small
%\resizebox{\columnwidth}{!}{
\begin{tabular}{ |c|c|c|c|c| }
\hline
Dysthymia & cfs & ibs & Narcolepsy & bulimia\\
\hline
Hypothyroidism & Costochondritis & psychosis & CysticFibrosis & POTS\\
\hline
MultipleSclerosis & Gastroparesis & gout & adhd & diabetes\\
\hline
CrohnsDisease & lupus & rheumatoid & Sinusitis & thyroidcancer\\
\hline
Hyperhidrosis & gerd & AnkylosingSpondylitis & endometriosis & schizophrenia\\
\hline
asthma & bipolar & depression & pcos & covid19\\
\hline
acne & anxiety & dementia & ptsd & dystonia\\
\hline
Epilepsy & ErectileDysfunction & Herpes & insomnia & Anemia\\
\hline
LymeDisease & migraine & ocd & parkinsons & Alzheimers\\
\hline
hpv & Prostatitis & backpain & Sciatica & Fibromyalgia\\
\hline
bpd & PagetSchroetter & & \\
\hline
\end{tabular}%}
%\vspace{-2mm}
\caption{\footnotesize A list of all 52 health conditions used in {\DatasetName}.}
%\vspace{-1mm}
\label{tab:all_health_conditions}
\end{table*}

\section{{\DatasetName} Annotation Details}
\label{appendix:human_dataset_details}

Human annotation for {\DatasetName} was done via paid crowdworkers on AMT, who were from Anglophone countries. They were selected through a series of qualification tests on a small subset of the samples, and have a history of high approval rates ($>95$\%) and good performance on related tasks. Based on initial annotations and performance on the qualification tests, workers were only re-hired if their performance was sufficient over time and they reliably followed the given instructions. The annotators were paid variable amounts (with periodic bonuses over time) depending on their performance and consistency, and the pay for all workers exceeds the minimum wage for the USA. %We neither solicit, record, request, or predict any personal information pertaining to the AMT crowdworkers.

The workers were asked to write passages (that include explanations) that are as specific and factually accurate as possible, describing how a specific dimension attribute relates to the given condition. Each HIT (annotation page) contains a single condition + dimension attribute combination, and they write a single passage that fills in the given template with an explanation. In the instructions, we describe each dimension in detail, and include several examples of correct and incorrect passages (regarding medical/factual accuracy, brevity/readability, and informativeness). We also encourage them to consult useful and trusted clinical resources such as MayoClinic, CDC, WebMD, and Healthline, if necessary, while writing the explanation.

Annotations were manually examined by the authors as they came in, and annotators were asked to improve their explanations if necessary. Annotators with consistently poor annotations were asked to stop annotating, and their completed annotations were re-annotated by others. At the end of the data collection process, the authors manually examined a large subset of {\DatasetName}, ensuring sufficiently high quality of annotations in terms of medical accuracy, informativeness, and readability.

%HTML templates which include instructions and examples can be found in the submitted supplementary .zip files.%a question snippet can be seen in Figure \ref{fig:human_eval_template}.

%\section{More Examples from {\DatasetName}}
%\label{appendix:more_dataset_examples}

\section{Further Model Finetuning and Generation Details}
\label{appendix:model_finetuning_generation}

T5-large consists of 770M params, T5-base 220M params, BART-large 406M params, and BART-base 139M params. For all models, we use beam search with a beam size of 5, decoder early stopping, a decoder length penalty of 0.6, and a decoder minimum length of 1. We set the maximum encoder and decoder lengths depending on values that can fit all examples in {\DatasetName}$_{tr}$, which ended up being 32 and 128 (for encoder and decoder, respectively) for the BART models, and 35 and 128 for the T5 models. Models are trained using fp16, and Adam optimizer with epsilon=1e-08. We use a training seed of 42 for all models, and a random seed of 42 for all other scripts that involved randomization. Decoding is done using beam search with a beam width of 5.

For model training, we use a batch size of either 64 or 32 for T5-base and BART-base, and either 8 or 16 for BART and T5-large (depending on GPU memory). For T5-base and BART-base, we use 400 warmup steps, 500 for BART-large, and 1200 for T5-large. We train all models up to a reasonable number of epochs (e.g. 20 to 30 for base models and 10 to 15 for large models). The learning rates for {\TaskName} models were determined by trying a range of values (e.g. from 1e-8 to 5e-1), and finding ones which led to good convergence behavior. %(e.g. validation metrics increase at a decently steady rate and reach max. after a reasonable number of epochs).
For the best-performing models, learning rates are as follows: BART-base = 5e-06, BART-large = 1e-05, T5-base = 1e-03, T5-large = 1e-05.

Training was done using single GTX 1080 Ti, TITAN RTX, RTX 2080 Ti, and GTX TITAN X GPUs. Model training time varied depending on the model type+size and amount of data augmentation, and varied between 5 minutes to 3 hours.

%Both are trained with generally the same hyperparameters as the baseline T5-base reimplementation, including decoding beam size of 5, but with a lower learning rate (LR) of 1e-5 as we found that higher LRs result in divergence.

%and T5-base models trained in approx. 1 hour, BART-base models in approx. 45 minutes, T5-large models in approx. 4 hours, and BART-large models in approx. 1.5-2 hours.

\section{Further Human Evaluation Details}
\label{appendix:human_eval_details}

Human evaluation was done via paid crowdworkers on AMT, who were from Anglophone countries. They were selected through qualification tests and have a history of high approval rates ($>95$\%) and good performance on related tasks. Each example was evaluated by 2 annotators. The time given for each AMT task instance or HIT was 1 hour maximum for an approximately 1-minute task. Sufficient time to read instructions, as calibrated by authors, was also considered. Annotators were paid 20 cents per HIT. This rate (\$12/hr) exceeds the minimum wage for the USA (\$7.25/hr) and constitutes fair pay. Workers who performed well were also paid periodic bonuses based on the timeliness and quality of their annotations. %We neither solicit, record, request, or predict any personal information pertaining to the AMT crowdworkers.

The human evaluation was split into 9 studies: 3 pairwise method comparisons ({\retr} vs. T5-large, {\retr} vs. human, and human vs. T5-large) by 3 dimensions (risk factors, treatment, and prevention). Each HIT or task page presented a given condition, the associated dimension attribute, and two explanations (from the two methods of the study) in a random order. They are asked to select among the three choices (first explanation is better, second explanation is better, \textit{hard to prefer one over the other}) for the three evaluation aspects of medical accuracy, informativeness, and readability. In the instructions, we describe the clinical dimension and each evaluation aspect in detail with positive and negative examples of each, and encouraged them to consult useful and trusted clinical resources such as MayoClinic, CDC, WebMD, and Healthline, if necessary, while evaluating the explanations.% HTML templates which include instructions can be found in the submitted supplementary .zip files. %a question snippet can be seen in Figure \ref{fig:human_eval_template}.

\begin{comment}
\begin{figure*}[!ht]
\begin{subfigure}{.97\textwidth}
    \centering
    \includegraphics[width=0.99\textwidth]{images/final_amt_instructions.png}
    \caption{}
    \label{fig:amt_preserve_instructions_graph}
\end{subfigure} \\
\begin{subfigure}{.97\textwidth}
    \centering
    \includegraphics[width=0.99\textwidth]{images/final_amt_questions_FINAL.png}
    \caption{}
    \label{fig:amt_preserve_questions_graph}
\end{subfigure}
\vspace{-0.8\abovedisplayskip}
%\vspace{-2mm}
\vspace{1mm}
\caption{\small Snapshots of human evaluation: a) instructions seen by annotator and b) an example with questions.\label{fig:human_eval_template}}
%\vspace{-5mm}
\end{figure*}
\end{comment}

\section{Further Qualitative Examples}
\label{appendix:qual-examples}

See Tables \ref{tab:qualitative_appendix_1} and \ref{tab:qualitative_appendix_2} for more qualitative examples on {\DatasetName}$_{test}$.

\begin{table*}[h!]
\centering
\small
%\scriptsize
%\begin{center}
%\scriptsize
\addtolength{\tabcolsep}{-4pt}
%\resizebox{\columnwidth}{!}{
\begin{tabular}{|p{1.2cm}|p{14.3cm}|}
\hline 
\textbf{Method} & \textbf{Text} \\ \hline
Input & A person with Hyperhidrosis has a/an infection risk factor because/since/as \{explanation\}\\\hline
Human & {\textcolor{teal}{A person with Hyperhidrosis has an infection risk factor because some kinds of infections like tuberculosis, HIV, bone infection cause hyperhidrosis}}\\\hline
{\retr} & \textcolor{red}{A person with Hyperhidrosis has a/an infection risk factor because/since/as this increases the risk of skin conditions such as athlete's foot and more severe conditions such as bacterial infections or pitted keratolysis. Study results report a nearly 30\% greater risk of skin infections in patients with hyperhidrosis compared with healthy controls.} \\ \hline
T5-L & {\textcolor{blue}{A person with Hyperhidrosis has an infection risk factor because infections such as staph, typhoid, hepatitis and polio can cause excessive sweating.}} \\ \hline 
T5-B & {\textcolor{brown}{ A person with Hyperhidrosis has an infection risk factor because a person who has a viral infection is more likely to have a virus infection.}} \\ \hline
BART-L & {\textcolor{violet}{A person with Hyperhidrosis has an infection risk factor because infection can cause hyperhidrosis}} \\ \hline 
BART-B & {\textcolor{orange}{A person with Hyperhidrosis has an infection risk factor because a person with hyperhidrosis is more likely to have an infection.}} \\
\Xhline{3\arrayrulewidth}
Input & A person with AnkylosingSpondylitis has a/an age risk factor because/since/as \{explanation\}\\\hline
Human & {\textcolor{teal}{A person with AnkylosingSpondylitis has an age risk factor because about 80 percent of people first develop ankylosing spondylitis symptoms under age 30}}\\\hline
{\retr} & \textcolor{red}{A person with AnkylosingSpondylitis has a/an age risk factor because/since/as about 80 percent of people first develop ankylosing spondylitis symptoms under age 30} \\ \hline
T5-L & {\textcolor{blue}{A person with AnkylosingSpondylitis has an age risk factor because people over 50 years of age are at a higher risk of developing the disease}} \\ \hline 
T5-B & {\textcolor{brown}{A person with AnkylosingSpondylitis has an age risk factor because older adults are more likely to develop ankylosis.}} \\ \hline
BART-L & {\textcolor{violet}{A person with AnkylosingSpondylitis has an age risk factor because the risk of developing ankylosis increases with age}} \\ \hline 
BART-B & {\textcolor{orange}{A person with AnkylosingSpondylitis has an age risk factor because older people are more likely to develop the disease}} \\
\Xhline{3\arrayrulewidth}
Input & A person with rheumatoid has a/an regular exercise treatment because/since/as \{explanation\}\\\hline
Human & {\textcolor{teal}{A person with rheumatoid has a/an regular exercise treatment because it helps reduce the disability often associated with rheumatoid arthritis and boosts strength and flexibility}}\\\hline
{\retr} & \textcolor{red}{A person with rheumatoid has a/an regular exercise treatment because/since/as but exercise is one of the key treatments to help reduce the disability often associated with rheumatoid arthritis. Regular exercise can boost strength and flexibility in people who have rheumatoid arthritis. Stronger muscles can better support your joints, while improved flexibility can aid joint function.} \\ \hline
T5-L & {\textcolor{blue}{A person with rheumatoid has a regular exercise treatment because exercise can help reduce inflammation and improve blood flow to the joints.}} \\ \hline 
T5-B & {\textcolor{brown}{A person with rheumatoid has a regular exercise treatment because regular exercise can help reduce the symptoms of the disease.}} \\ \hline
BART-L & {\textcolor{violet}{A person with rheumatoid has a regular exercise treatment because exercise can help reduce inflammation and pain}} \\ \hline 
BART-B & {\textcolor{orange}{A person with rheumatoid has a regular exercise treatment because regular exercise can help reduce inflammation and inflammation in the body.}} \\
\hline
\end{tabular}%}
%\vspace{-2mm}
\caption{\footnotesize More qual examples for{\DatasetName}$_{test}$: \textcolor{teal}{human}, \textcolor{red}{{\retr}}, \textcolor{blue}{T5-large (L)}, \textcolor{brown}{T5-base (B)}, \textcolor{violet}{BART-large (L)}, \textcolor{orange}{BART-base (B)}}
\label{tab:qualitative_appendix_1}
%\vspace{-2mm}
\end{table*}
%\vspace*{-2mm}

\begin{table*}[h!]
\centering
\small
%\scriptsize
%\begin{center}
%\scriptsize
\addtolength{\tabcolsep}{-4pt}
%\resizebox{\columnwidth}{!}{
\begin{tabular}{|p{1.2cm}|p{14.3cm}|}
\hline 
\textbf{Method} & \textbf{Text} \\ \hline
Input & A person with thyroidcancer has a/an chemotherapy treatment because/since/as \{explanation\}\\\hline
Human & {\textcolor{teal}{A person with thyroidcancer has a chemotherapy treatment because Chemotherapy (chemo) uses anti-cancer drugs that are injected into a vein or are taken by mouth. Chemotherapy is systemic therapy, which means that the drug enters the bloodstream and travels throughout the body to reach and destroy cancer cells and stops cancer growth.}}\\\hline
{\retr} & \textcolor{red}{A person with thyroidcancer has a/an chemotherapy treatment because/since/as chemotherapy is seldom helpful for most types of thyroid cancer, but fortunately it is not needed in most cases. It is often combined with external beam radiation therapy for anaplastic thyroid cancer and is sometimes used for other advanced cancers that no longer respond to other treatments.} \\ \hline
T5-L & {\textcolor{blue}{A person with thyroidcancer has a chemotherapy treatment because chemotherapy destroys the cells that cause the cancer.}} \\ \hline 
T5-B & {\textcolor{brown}{A person with thyroidcancer has a chemotherapy treatment because chemotherapy reduces the risk of thyroid cancer.}} \\ \hline
BART-L & {\textcolor{violet}{A person with thyroidcancer has a chemotherapy treatment because chemotherapy can slow down the growth of thyroid cancer}} \\ \hline 
BART-B & {\textcolor{orange}{A person with thyroidcancer has a chemotherapy treatment because chemotherapy can help reduce the risk of thyroid cancer}} \\
\Xhline{3\arrayrulewidth}
Input & A person with depression has a/an medical conditions risk factor because/since/as \{explanation\}\\\hline
Human & {\textcolor{teal}{A person with depression has a medical conditions risk factor because the pain and stress that come with certain conditions, including cancer, stroke, or heart disease, can take a toll on a person's mental state resulting in depression.}}\\\hline
{\retr} & \textcolor{red}{A person with depression has a/an medical conditions risk factor because/since/as more commonly studied were the connections between depression and cardiovascular disease, metabolic syndrome, biochemical alterations, diabetes, dementia, cognitive impairment, Alzheimer's disease, somatization and chronic pain, asthma, arthritis, and hyperlipidemia.} \\ \hline
T5-L & {\textcolor{blue}{A person with depression has a medical conditions risk factor because depression can be caused by a number of medical conditions, such as diabetes, heart disease, high blood pressure, high cholesterol and high blood sugar.}} \\ \hline 
T5-B & {\textcolor{brown}{A person with depression has a medical conditions risk factor because medical conditions such as diabetes, heart disease, diabetes, AIDS, and other conditions that cause depression, such as cancer, are linked to depression.}} \\ \hline
BART-L & {\textcolor{violet}{A person with depression has a medical conditions risk factor because depression can be caused by a variety of medical conditions.}} \\ \hline 
BART-B & {\textcolor{orange}{A person with depression has a medical conditions risk factor because depression can be a cause of depression.}} \\
\Xhline{3\arrayrulewidth}
Input & A person with depression has a/an grief risk factor because/since/as \{explanation\}\\\hline
Human & {\textcolor{teal}{A person with depression has a grief risk factor because grief associated with loss of dear one greatly increases the risk of psychiatric complications such as depression}}\\\hline
{\retr} & \textcolor{red}{A person with depression has a/an grief risk factor because/since/as risk Factors · Genetics: A history of depression in your family may make it more likely for you to get it. · Death or loss: Sadness and grief are ...} \\ \hline
T5-L & {\textcolor{blue}{A person with depression has a grief risk factor because people who have experienced a loss of a loved one are more likely to develop depression.}} \\ \hline 
T5-B & {\textcolor{brown}{A person with depression has a grief risk factor because grief is the most common cause of depression.}} \\ \hline
BART-L & {\textcolor{violet}{A person with depression has a grief risk factor because grief can cause depression}} \\ \hline 
BART-B & {\textcolor{orange}{A person with depression has a grief risk factor because grief is associated with depression.}} \\
\hline
\end{tabular}%}
%\vspace{-2mm}
\caption{\footnotesize More qual examples for{\DatasetName}$_{test}$: \textcolor{teal}{human}, \textcolor{red}{{\retr}}, \textcolor{blue}{T5-large (L)}, \textcolor{brown}{T5-base (B)}, \textcolor{violet}{BART-large (L)}, \textcolor{orange}{BART-base (B)}}
\label{tab:qualitative_appendix_2}
%\vspace{-2mm}
\end{table*}

\end{document}